\pdfoutput=1

\documentclass[11pt]{article}

\usepackage{EACL2023}

\usepackage{times}
\usepackage{latexsym}

\usepackage[T1]{fontenc}

\usepackage[utf8]{inputenc}

\usepackage{microtype}

\usepackage{inconsolata}

\usepackage{amsmath}
\usepackage{amssymb}
\usepackage{booktabs}
\usepackage{bbm}
\usepackage{graphicx}
\usepackage{geometry}
\usepackage[colorinlistoftodos,prependcaption,textsize=tiny]{todonotes}
\usepackage[smallerops,varbb]{newtxmath}
\usepackage{xspace}
\usepackage{float}
\usepackage{multirow}
\usepackage{nicefrac}
\usepackage[export]{adjustbox}

\usepackage{enumitem}
\DeclareMathAlphabet{\mathcal}{OMS}{cmsy}{m}{n}
\DeclareMathAlphabet{\mathbb}{U}{msb}{m}{n}
\setlist[itemize]{leftmargin=*}

\newcommand*{\courier}{\fontfamily{pcr}\selectfont}
\newcommand{\jhu}{\textsuperscript{\rm 1}}
\newcommand{\ur}{\textsuperscript{\rm 2}}
\newcommand{\mssm}{\textsuperscript{\rm 3}}

\title{Iterative Document-level Information Extraction via Imitation Learning}

\author{Yunmo Chen\jhu\quad William Gantt\ur\quad Weiwei Gu\ur \\
\bf Tongfei Chen\mssm\quad Aaron Steven White\ur\quad Benjamin Van Durme\jhu \\
  \jhu~Johns Hopkins University~~\ur~University of Rochester~~\mssm~Microsoft Semantic Machines \\
  \courier{\small\{yunmo|vandurme\}@jhu.edu, \{wgantt@ur.|wgu7@ur.|aaron.white@\}rochester.edu, tongfei@pm.me}
}

\newcommand{\iterx}{\textsc{IterX}\xspace}
\newcommand{\scirex}{\textsc{SciREX}\xspace}

\newcommand{\muc}{\textsc{MUC-4}\xspace}

\newcommand{\rmephi}{CEAF-RME\textsubscript{$\phi_3$}\xspace}
\newcommand{\reedef}{CEAF-REE\textsubscript{def}\xspace}
\newcommand{\reeimpl}{CEAF-REE\textsubscript{impl}\xspace}
\newcommand{\rmesubset}{CEAF-RME\textsubscript{$\phi_\subseteq$}\xspace}

\usepackage[scaled=0.85]{FiraMono}

\begin{document}

\maketitle
\begin{abstract}
We present a novel iterative extraction model, \iterx, for extracting complex relations, or \emph{templates} (i.e., $N$-tuples representing a mapping from named slots to spans of text) within a document. Documents may feature zero or more instances of a template of any given type, and the task of \emph{template extraction} entails \emph{identifying} the templates in a document and \emph{extracting} each template's slot values. Our imitation learning approach casts the problem as a Markov decision process (MDP), and relieves the need to use \emph{predefined template orders} to train an extractor. It leads to state-of-the-art results on two established benchmarks -- 4-ary relation extraction on \scirex and template extraction on \muc \,-- as well as a strong baseline on the new BETTER Granular task.\footnote{~Code available at \href{https://www.github.com/wanmok/iterx}{\tt github.com/wanmok/iterx}.}
\end{abstract}

\section{Introduction}\label{sec:intro}
A variety of tasks in information extraction (IE) require synthesizing information across multiple sentences, up to the length of an entire document. The centrality of document-level reasoning to IE has been underscored by an intense research focus in recent years on problems such as argument linking \citep[\emph{i.a.}]{ebner-etal-2020-multi, li-etal-2021-document}, $N$-ary relation extraction \citep[\emph{i.a.}]{quirk-poon-2017-distant, yao-etal-2019-docred, jain-etal-2020-scirex}, and --- our primary focus --- template extraction \citep[][\textit{i.a.}]{du-etal-2021-template, huang-etal-2021-document}.

\begin{figure}
    \centering
    \includegraphics[width=\columnwidth]{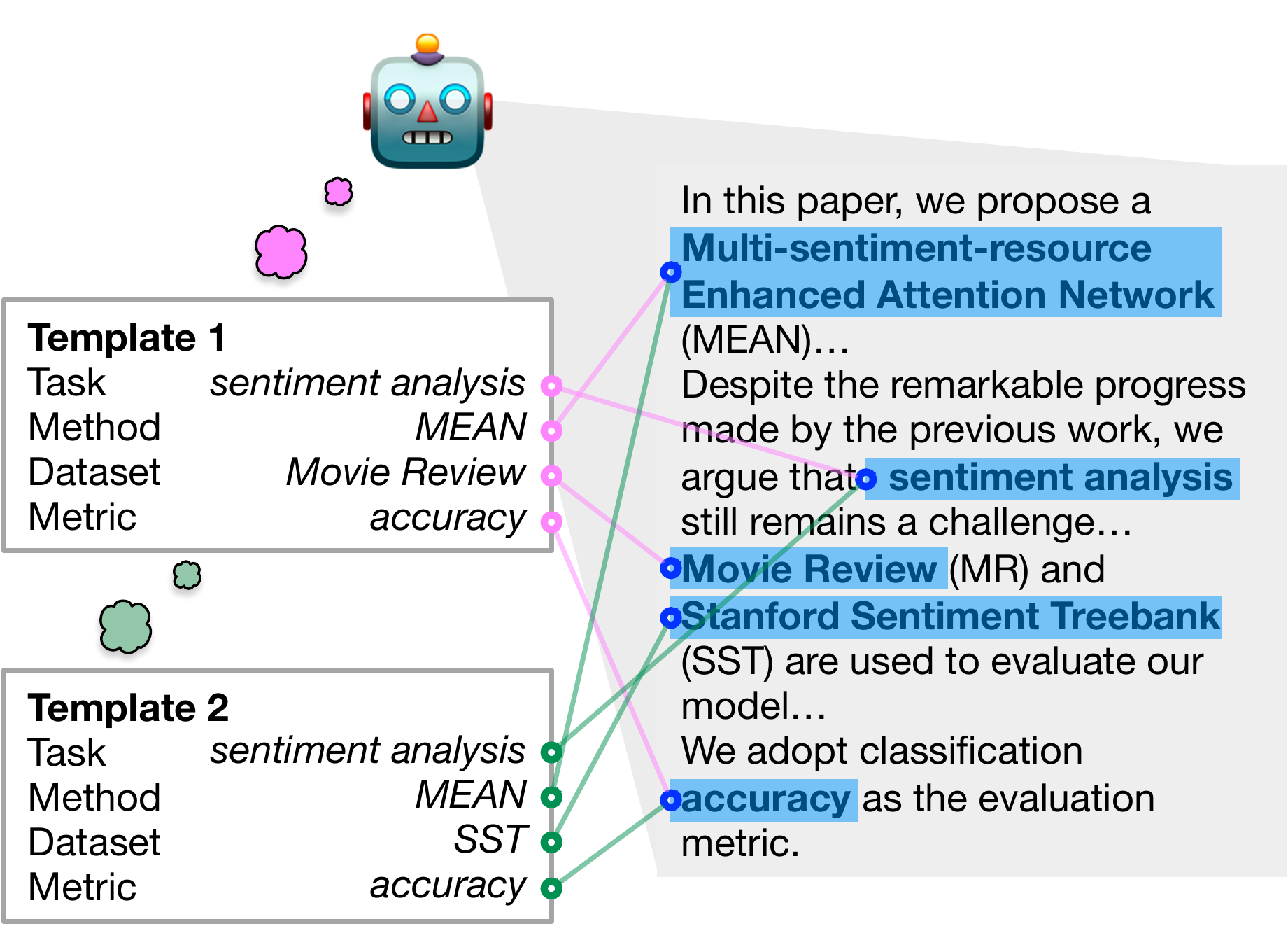}
    \caption{An example of multi-template extraction on a document (an NLP paper; \citet{lei-etal-2018-multi}) from the \scirex dataset. An agent reads the entire paper and \emph{iteratively} generates templates, each consisting of slots for \texttt{Task}, \texttt{Method}, \texttt{Dataset}, and \texttt{Metric}.}
    \label{fig:multi_template}
    \vspace{-5mm}
\end{figure}

Construed broadly, template extraction is general enough to subsume certain other document-level extraction tasks, including $N$-ary relation extraction. Motivated by this consideration, we propose to treat these problems under a unified framework of \emph{generalized template extraction} (\S\ref{sec:formulation}).\footnote{~We encourage the reader to consult Appendix \ref{app:terminology} for a discussion of some important differences between generalized template extraction and traditional event extraction. } \autoref{fig:multi_template} shows 4-ary relations from the \scirex dataset \cite{jain-etal-2020-scirex}, presented as simple templates.

\begin{figure*}[!ht]
    \centering
    \includegraphics[width=\textwidth]{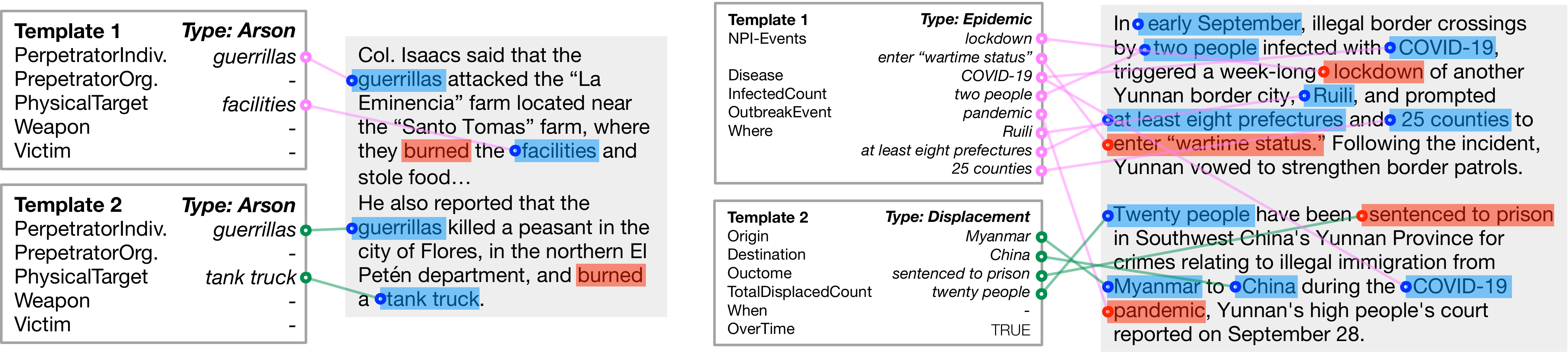}
    \caption{Template examples from \muc (left) and BETTER Granular (right) datasets. Event triggers (e.g. \textit{burned} above) are \emph{not} annotated in \muc\ and are highlighted here only for clarity.}
    \label{fig:muc-better-egs}
\end{figure*}

Since documents typically describe multiple complex events and relations, template extraction systems must be capable of predicting multiple templates per document. Existing approaches such as \citet{du-etal-2021-template} and \citet{huang-etal-2021-document} rely on a linearization strategy to force models to learn to predict templates in a \emph{pre-defined order}. In general, however, such orderings are  arbitrary. Others have instead focused on the simplified problem of role-filler entity extraction (REE), which entails extracting all slot-filling entities but does not involve mapping them to individual templates \citep[\textit{i.a.}]{patwardhan-riloff-2009-unified, du-etal-2021-grit}.

We present a new model for generalized template extraction, \iterx, that iteratively extracts multiple templates from a document \emph{without} requiring a pre-defined linearization scheme. We formulate the problem as a \emph{Markov decision process} (MDP, \S\ref{sec:formulation}), where an action corresponds to the generation of a single template (\S\ref{subsec:policy}), and states are sets of predicted templates (\S\ref{subsec:state-transition}). Our system is trained via \emph{imitation learning}, where the agent imitates a dynamic oracle drawn from an expert policy (\S\ref{subsec:training}). Our contributions can be summarized as follows:

\begin{itemize}\setlength\itemsep{-0.15em}
    \item We show that generalized template extraction can be treated as a Markov decision process, and that imitation learning can be effectively used to train a model to learn this process without making explicit assumptions about template orderings.
    \item We demonstrate state-of-the-art results with \iterx\ on two established benchmarks for complex relation extraction: 4-ary relation extraction on \scirex and template extraction on \muc.
    \item We introduce strong baselines for the recently introduced English BETTER Granular template extraction task.
\end{itemize}

\section{Problem Formulation}\label{sec:formulation}

We propose to treat both classic template extraction and $N$-ary relation extraction under a unified framework of \emph{generalized template extraction}. Given a document $D = (w_1, \cdots, w_N)$ where each $w_i$ is a token, we assume that some system (or model component) generates a set of candidate \emph{mention spans} $\mathcal{X} = \{x_1, \cdots, x_M\}$, where each $x_i = D[l_i : r_i] \in \mathcal{X}$ is contiguous with left and right span boundary indices $l_i$ and $r_i$.

We define a \emph{template ontology} as a set of template types $\mathcal{T}$, where each type $t\in\mathcal{T}$ is associated with a set of slot types $S_t$. A \textit{template instance} is defined as a pair $(t,\{(s_k: X_k), \cdots\})$ where $t \in \mathcal{T}$ is a template type, $s_k \in S_t$ is a \emph{slot type} associated with $t$, and $X_k \subseteq \mathcal{X}$ is a subset of all mention spans extracted from the document that fills slot type $s_k$ ($X_k = \varnothing$ indicates that slot $s_k$ has no filler).
\footnote{~In this work, we use \textit{template} as an abbreviation for \textit{template instance}, relying on the \textit{type} vs.\ \textit{instance} usage only when necessary for clarity.} 
 Taking \autoref{fig:muc-better-egs} (left) as an example, Template 1 has type $t = \texttt{Arson}$ and slots $\{\texttt{PerpretratorIndiv}: \{\textrm{``guerrillas''}\}, \texttt{PhysicalTarget}: \{\textrm{``facilities''}\}\}$.

We reduce the problem of extracting a \emph{single} template to the problem of assigning a slot type to each extracted span $x_i \in \mathcal{X}$, where some spans may be assigned a special \emph{null} type ($\varepsilon$), indicating that they fill no slot in the current template. Given this formulation, we can equivalently specify a template instance as $(t, a)$ where $a$ is an \emph{assignment} of spans to slot types: $\{x_i : s_i \}_{s_i \in S_t}$. We denote the union of all slot types across all template types, along with the empty slot type $\varepsilon$, as $\mathcal{S} = \{\varepsilon\} \cup {\displaystyle\bigcup}_{t\in\mathcal{T}}S_t$.

With these definitions in hand, the problem of \emph{generalized template extraction} can be stated succinctly: Given a template ontology $\mathcal{T}$, a document $D$, and a set of candidate mentions $\mathcal{X}$ extracted from $D$, generate a set of template instances $\{(t_1, a_1), \cdots, (t_K, a_K)\}$, where $t_i \in \mathcal{T}$.

\begin{figure*}[t!]
    \centering
    \includegraphics[width=1\textwidth]{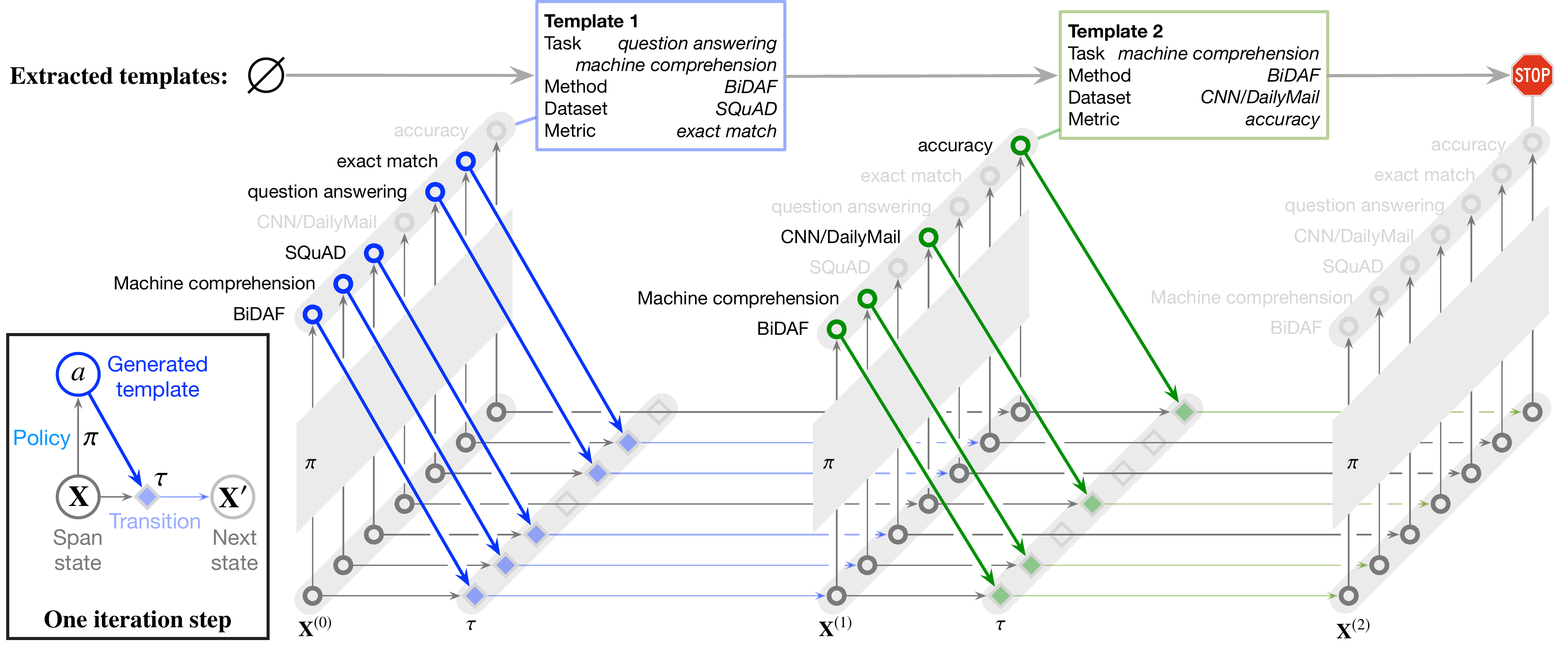}
    \caption{The basic iteration step of \iterx (left box), and an \textbf{unrolled} version on \scirex~4-ary relation extraction (extraction of templates in the form \{\texttt{Task}, \texttt{Method}, \texttt{Dataset}, \texttt{Metric}\}) executed on the NLP paper {\it Bidirectional Attention Flow for Machine Comprehension} \cite{SeoKFH17}. Span embeddings ({$\mathbf{X}^{(0)}$}) are passed as input to the first step, where the model extracts the template \{\texttt{Task}: \emph{Machine comprehension, Question answering}; \texttt{Method}: \emph{BiDAF}; \texttt{Dataset}: \emph{SQuAD}; \texttt{Metric}: \emph{Exact match}\}. This information is propagated via our \textit{memory mechanism} to the second step, and informs prediction of the next template: \{\texttt{Task}: \emph{Machine comprehension}; \texttt{Method}: \emph{BiDAF}; \texttt{Dataset}: \emph{CNN/DailyMail}; \texttt{Metric}: \emph{Accuracy}\}. The third step assigns the null slot type $\varepsilon$ to all spans, indicating that the model is unable to find any further templates, thus stopping the generation process.}
    \label{fig:iterx-arch}
\end{figure*}

\paragraph{As an MDP}
We treat generalized template extraction as a Markov decision process (MDP), where each step of the process generates one whole template instance. For simplicity, we consider the problem of extracting templates of a specific type $t \in \mathcal{T}$; extracting \emph{all} templates then simply requires iterating over $\mathcal{T}$, where $|\mathcal{T}|$ is typically very small. This MDP $(2^\mathcal{A}, \mathcal{A}, E, R)$
comprises the following:\footnote{~Our notation is consistent with prior NLP work that uses MDPs, e.g., Levenshtein Transformers \citep{GuWZ19}.} 
\begin{itemize}\setlength\itemsep{-0.05em}
    \item $2^\mathcal{A}$: the set of states. In our case, this is the set of all template generation histories. Each state $A \subseteq \mathcal{A}$ is a set of generated templates;
    \item $\mathcal{A}$: the set of \emph{actions} or \emph{assignments}: an action is the generation of a single template (an assignment of slot types to spans);
    \item $E$: the environment that dictates state transitions. Here, each transition simply adds a generated template to the set of all templates generated for the document: $E(A, a) = A \cup \{a\}$;
    \item $R(A, a)$: the reward from action $a$ under the current state $A$.
\end{itemize}
 These components are detailed in the following section. \autoref{fig:iterx-arch} shows \iterx in action: the MDP produces two templates sequentially, terminating on a null assignment to all input spans in $\mathcal{X}$.

\section{Model}\label{sec:model}

Our \iterx model is a parameterized agent that makes decisions under the MDP above: conditioned on an input document $D$, spans $\mathcal{X}$ extracted from $D$, and a specific template type $t$, \iterx\ generates a single template of type $t$ at each step. The model consists  of two parameterized components:
\begin{itemize}\setlength\itemsep{-0.05em}
    \item \textbf{Policy $\boldsymbol{\pi}$}: A policy $\pi(a \mid t, \mathbf{X})$ that generates a distribution over potential assignments of spans to slots in the current template of type $t$;
    \item \textbf{State transition model $\boldsymbol{\tau}$}: An autoregressive state encoder that maps a state (i.e., a set of predicted templates) $A$ to a continuous representation $\mathbf{X}$ via a state transition model $\tau$, where $\mathbf{X}^{(k+1)} = \tau(\mathbf{X}^{(k)}, a_k)$. Here the state representation $\mathbf{X} = (\mathbf{x}_1, \cdots, \mathbf{x}_M)$ comprises of a vector $\mathbf{x}_i \in \mathbb{R}^d$ for each span $x_i$.
\end{itemize}

\noindent \iterx generates a sequence of templates: It starts with initial state $\mathbf{X}^{(0)}$ (see \autoref{fig:iterx-arch} for a running example) comprising only span representations derived from the encoder (as no templates have been predicted) and ends when no new template is generated (\S\ref{subsec:inference}). \iterx is trained via imitation learning, aiming to imitate an expert policy $\pi^*$ derived from reference templates.

\subsection{Span Extraction and Representation}\label{subsec:span-extraction}
\iterx\ takes \textit{spans} as input and thus relies on a span proposal component to obtain candidate spans.\footnote{~Although many systems since \citet{lee-etal-2017-end} that require spans as input have trained span proposal modules end-to-end, we found this to be unnecessary to obtain strong results and leave this extension for future work.} 
For all experiments, we use the neural CRF-based span finder employed for FrameNet parsing in \citet{xia-etal-2021-lome} and for the BETTER Abstract task in \citet{yarmohammadi-etal-2021-everything}.\footnote{~The code of the span finder can be found at \href{https://github.com/hiaoxui/span-finder}{\tt github.com/hiaoxui/span-finder}. We refer the reader to these papers for further details.} CRF-based span finders have been empirically shown to excel at IE tasks \cite{spanfinding}.

The input document is first embedded using a pretrained Transformer encoder~\citep{devlin-etal-2019-bert,RaffelSRLNMZLL20} that is fine-tuned during model training.\footnote{~Different encoders are used in \S\ref{sec:experiments} for fair comparison with prior work. For documents exceeding maximum length $N_{\max}=1024$, tokens are encoded in chunks of size $N_{\max}$.} Each span $x = D[l: r]$ extracted by the span extractor is encoded as a vector $\mathbf{x}_{\rm enc}$, which is obtained by first concatenating three vectors of dimension $d$: the embeddings of the first and the last tokens in the span, and a weighted sum of the embeddings of all tokens within the span, using a learned global query~\citep{lee-etal-2017-end}. This $3d$-dimensional vector is then compressed to size $d$ using a two-layer feedforward network. Lastly, to incorporate positional information, we add sinusoidal positional embeddings based on the token-level offset of $l$ within the document to yield $\mathbf{x}_{\rm enc}$.

\subsection{Policy: Generating a Single Template}\label{subsec:policy}
A policy $\pi$ generates a single template given span states $\mathbf{X} = (\mathbf{x}_1, \cdots, \mathbf{x}_n)$ and template type $t \in \mathcal{T}$, conditioned on the document and all of its candidate mention spans.

Since an action $a$ represents a set of \emph{slot type assignments} for all candidate mentions, the policy $\pi(a \mid t, \mathbf{X})$ can be factorized as
\begin{equation}
    \pi(a \mid t, \mathbf{X}) = \prod_{(x : s) \in a} P(s|t, \mathbf{x}) \ .
\end{equation}

\noindent Thus we only need to model the slot type distribution for each candidate span. Here, we employ two models described below. 

\paragraph{Independent Modeling} We train a classifier that outputs a slot type (or $\varepsilon$) given both the template type embedding $\mathbf{t}$ and the slot type embedding $\mathbf{s}$, inspired by a standard practice in binary relation extraction \citep[\emph{i.a.}]{ebner-etal-2020-multi, lin-etal-2020-joint}. It computes the probability with a two-layer feedforward network (FFN), with slots not associated with the template type (i.e., $s \not\in S_t \cup  \{\varepsilon\})$ assigned 0 probability:
\begin{equation}\label{eq:indep-policy}
    P_{\rm ind}(s|t, \mathbf{x}) \propto \mathbf{1}_{s \in S_t \cup \{\varepsilon\}} \cdot \exp (\mathbf{s}^{\rm T} \cdot \mathrm{FFN}([\mathbf{t}; \mathbf{x}]))
\end{equation}
where $[\cdot ~;~ \cdot]$ denotes vector concatenation.

\paragraph{Joint Modeling}
Following \citet{chen-etal-2020-joint-modeling}, we create a model that \textit{jointly} considers all candidate spans given the template type. We begin by prepending $\mathbf{t}$ to the sequence of span states $\mathbf{X}$ to yield the sequence $(\mathbf{t}, \mathbf{x}_1, \cdots, \mathbf{x}_n)$. This sequence is fed to a different Transformer encoder, which naturally models interactions both between spans and between a span and the template type via self-attention \cite{vaswani2017attention}:
\begin{equation}\label{eq:tie}
(\hat{\mathbf{t}}, \hat{\mathbf{x}}_1, \cdots\!, \hat{\mathbf{x}}_M) = \mathrm{Transformer}(\mathbf{t}, \mathbf{x}_1, \cdots\!, \mathbf{x}_M)
\end{equation}
We emphasize that the inputs to the Transformer are embeddings of \emph{spans} (see \S\ref{subsec:span-extraction}) and not tokens, following \citet{chen-etal-2019-improving-long,chen-etal-2020-joint-modeling}.\footnote{~Positional embeddings are not needed in this Transformer since sinusoidal embeddings are already added to the span representations.}
For each $\mathbf{x}_i$, we pass the representation $\hat{\mathbf{ x}}_i$ output by the Transformer to a linear layer with output size $|\mathcal{S}|$, the total number of slot types. A softmax activation is then applied over all slot types $s$ that are valid for $t$ (i.e., $s \in S_t \cup  \{\varepsilon\})$, with invalid types masked out, yielding the following distribution:
\begin{equation}\label{eq:joint-policy}
    P_{\rm joint}(s|t, \mathbf{x}) \propto \mathbf{1}_{s \in S_t \cup \{\varepsilon\}}  \cdot \exp (\mathbf{s}^{\rm T} \hat{\mathbf{x}})
\end{equation}

\subsection{State Transition Model}\label{subsec:state-transition}
A \emph{state transition model} models the environment $E(A, a)$. Recall that a state transition just consists in the generation of a single template, where the current state $A$ is the set of all templates that have been generated up to the current step.

Here, we propose a neural model that produces a representation of $A$. Specifically, we model $A$ as a sequence of vectors $\mathbf{X}_{\rm mem}(A) \in \mathbb{R}^{M \times d}$ --- one $d$-dimensional state vector for each of the $M$ candidate spans $x \in \mathcal{X}$. Each state vector ${\mathbf{x}_{\rm mem}} \in \mathbb{R}^d$ acts as a \emph{span memory}, tracking the \textit{use} of that span across generated templates. We model state transitions using a single gated recurrent unit \citep[GRU; ][]{cho-etal-2014-properties}. Given the current template assignment $(x: s) \in a$ of a slot type $s$ to a span $x$, the state transition for $A^\prime = A \cup \{a\}$ is given as follows:
\begin{equation}
     \mathbf{x}_{\rm mem}^\prime = 
    \begin{cases}
      \mathrm{GRU}(\mathbf{x}_{\rm mem}, [ \mathbf{s}~;~\hat{\mathbf{t}}]) & \text{if~} s \ne \varepsilon; \\
      \mathbf{x}_{\rm mem} & \text{if~} s = \varepsilon.
    \end{cases}
\end{equation}
where $\hat{\mathbf{t}}$ is a template embedding given by $\hat{\mathbf{t}} = \mathbf{t}$ when using the independent policy model given in \autoref{eq:indep-policy} and given by \autoref{eq:tie} when using the joint model. Intuitively, $\hat{\mathbf{t}}$ is a summarized vector of the current template, akin to the role of the [\textsc{cls}] token employed in BERT \cite{devlin-etal-2019-bert}. Here, we use a concatenation of the slot type embedding $\mathbf{s}$ and the template vector $\hat{\mathbf{t}}$ as the input to the state transition GRU to track the use of the span.

The input representation of a span $x$ at each step is simply $\mathbf{x} = \mathbf{x}_{\rm enc} + \mathbf{x}_{\rm mem}$ --- the sum of the original span embeddings $\mathbf{x}_{\rm enc}$ described in \S\ref{subsec:span-extraction} and the current memory vector $\mathbf{x}_{\rm mem}$.

\subsection{Policy Learning}\label{subsec:training}

We use \emph{direct policy learning} (DPL), a type of imitation learning, to train our model. DPL entails training an agent to imitate the behavior of an \emph{interactive demonstrator} as given by optimal actions $a^*$ drawn from some expert policy $\pi^*$, a proposal distribution over actions. This expert policy is computed dynamically based on the current state of the agent, as we describe below. For this reason, the interactive demonstrator is sometimes referred to as a \textit{dynamic oracle} \citep{goldberg-nivre-2012-dynamic}.

\emph{The log-likelihood of the oracle action under the \iterx policy model is the reward}.

This ensures that the learning problem can be optimized directly using gradient descent, where the objective is given by the expected reward:

\begin{equation}\label{eq:mix-policy}
    \mathbb{E}_{\substack{a^* \sim \pi^* \\ A \sim d_{\tilde\pi}}} \left[ \sum_{k=0}^{\infty} \gamma^k \log \pi(a^* | t, \mathbf{X}(A)) \right]
\end{equation}

\noindent Here, $\gamma$ is a discount factor, $\tilde\pi$ is the \emph{mixed} policy, and states are repeatedly sampled from their induced state distribution $d_{\tilde\pi}$.  The mixed policy $\tilde\pi$ is a mixture of the expert policy and the agent's policy \cite{RossGB11}. Sampling from $\tilde\pi$ can thus be described as first sampling some $u \in \{0, 1\}$, then sampling from the agent's parameterized policy $\pi$ if $u = 1$, or sampling an action from the dynamic oracle $\pi^*$ if $u = 0$:
\begin{align}
    u     & \sim  \mathrm{Bernoulli}(\alpha); \nonumber \\
    \hat a& \sim  u\pi + (1-u)\pi^*.
\end{align}

\noindent Here $\alpha$, the \emph{agent roll-out rate}, or the \emph{agent policy mixing rate}, is a hyperparameter that controls the probability of the agent following its own policy vs. the dynamic oracle.

This process resembles \emph{scheduled sampling} \cite{BengioVJS15}, a technique commonly employed in training models for sequence generation tasks like machine translation: when updating decoder hidden states, either the gold token $y^*$ or the predicted token $\hat y$ may be used, and the decision is made via a random draw. Here, the difference is that we are generating templates at each step instead of tokens.

\paragraph{Expert Policy}
We construct an expert policy based on the agent's policy. At training time, given the set of gold templates $A^*$ and the current state $A$ (all templates predicted thus far), the set $\bar A = A^* \setminus A$ contains \emph{all gold templates not yet predicted}. Our expert policy is formulated as
\begin{equation}\label{eq:expert-policy}
    \pi^*(a \mid t, \mathbf{X}) \propto 
    \begin{cases}
      e^{\log \pi(a \mid t, \mathbf{X}) / \beta} & \text{if~} a \in \bar A \\
      0 & \text{if~} a \notin \bar A
    \end{cases}~,
\end{equation}
where $\beta$ is a temperature parameter.

Intuitively, our expert policy seeks to ``please'' the agent: a (viable) action's probability under the expert policy is proportional to the probability under the agent's policy. Temperature $\beta$ controls concentration: $\beta \to 0^+$ reduces it to a point distribution over a single action and $\beta \to \infty$ results in equal probability assigned to all remaining gold templates.

\subsection{Inference}\label{subsec:inference}
Although many search algorithms for sequence prediction can be employed (e.g. beam search, A*), we find greedy decoding to be effective, and leave further exploration for future work. Setting the initial state $A^{(0)} = \varnothing$, we take actions (i.e., generate templates) by greedy decoding $\hat a = \arg\max_a \pi(a \mid t, \mathbf{X})$ for every step. Decoding stops when all spans are assigned the null slot type $\varepsilon$ in $\hat a$.

\section{Experiments}\label{sec:experiments}

We evaluate \iterx on three datasets: \scirex \cite{jain-etal-2020-scirex}, \muc \cite{grishman-sundheim-1996-message}, and BETTER Phase II English Granular.\footnote{~\url{https://ir.nist.gov/better}.} \scirex is a challenge dataset for 4-ary relation extraction\footnote{~\scirex also contains a binary relation extraction task, but the binary relation is a subrelation of the 4-ary relation, and thus is subsumed by the more difficult task.} on full academic articles related to machine learning. \muc and Granular are both traditional template extraction tasks, though they differ in  important respects, which we discuss in \autoref{app:datasets}. For summary statistics, see \autoref{tab:datasets}.

\begin{table}[h]
    \centering
    \adjustbox{max width=\linewidth}{
    \setlength{\tabcolsep}{3pt}
    \begin{tabular}{lccccccccc}
        \toprule
        & \multicolumn{3}{c}{\bf \scirex} & \multicolumn{3}{c}{\bf MUC-4} & \multicolumn{3}{c}{\bf Granular} \\
        \cmidrule(lr){2-4} \cmidrule(lr){5-7} \cmidrule(lr){8-10}
        & Train & Dev & Test & Train & Dev & Test & Train & Dev & Test \\
        \midrule
        \# documents & 306 & 66 & 66 & 1,300 & 200 & 200 & 302 & 34 & 32 \\
        \# templates & 1,627 & 251 & 271 & 1,114 & 191 & 209 & 610 & 57 & 47 \\
        \midrule \midrule
        \# temp. types & \multicolumn{3}{c}{1} & \multicolumn{3}{c}{6} & \multicolumn{3}{c}{6} \\
        \# slot types & \multicolumn{3}{c}{4} & \multicolumn{3}{c}{5} & \multicolumn{3}{c}{92 + 4\textsuperscript{\dag}}\\
        \bottomrule

    \end{tabular}
    }
    \caption{Summary statistics of the datasets.  \textsuperscript{\dag} indicates slot types that take non-span values as fillers.}
    \label{tab:datasets}
\end{table}

\subsection{Baselines}\label{subsec:baselines}

\paragraph{\textsc{GTT} \citep{du-etal-2021-template}}
To our knowledge, this is the only prior work to have attempted full template extraction in recent years, and it is thus our primary baseline for comparison on \muc.\footnote{~We were unfortunately unable to obtain reasonable performance with GTT on \scirex, and so do not compare \iterx and GTT on this task.} 
GTT first prepends the document text with the valid template types, then passes the result to a BERT encoder. A Transformer decoder (whose parameters are shared with the encoder) then generates a linearized sequence of template instances.

\paragraph{\textsc{TempGen} \citep{huang-etal-2021-document}}
This is the current state-of-the-art system for REE (the simplified slot-filling entity extraction task) on \muc. On \scirex, \textsc{TempGen} may output multiple relation instances, but only one canonical mention as the filler for each role in the relation. On \muc, \textsc{TempGen} outputs a single aggregate template per document, but allows multiple spans to fill a template slot. We make minimal modifications to the \textsc{TempGen} source code to support multi-filler, multi-template prediction on both datasets, allowing for direct comparison to \iterx and GTT on full template extraction.

\subsection{Metrics}\label{subsec:exp-metrics}

The currently used metric for template extraction and REE on \muc is CEAF-REE, proposed in \citet{du-etal-2021-grit} and then used in \citet{du-etal-2021-template} and \citet{huang-etal-2021-document}.\footnote{~The standard metrics for template extraction may be unfamiliar to IE researchers more accustomed to sentence-level \textit{event extraction}. Accordingly, we thoroughly motivate and describe all template extraction metrics we use in this work in Appendix \ref{app:evaluation}. What follows is a more abridged discussion.} CEAF-REE is based on the CEAF metric \citep{luo-2005-coreference} for coreference resolution, that computes an alignment between gold and predicted entities that maximizes a measure of similarity $\phi$ between aligned entities (e.g.\ $\rm CEAF_{\phi_4}$ in coreference resolution). This alignment is subject to the constraint that each reference entity is aligned to at most one predicted entity.

The CEAF-REE \emph{implementation} (henceforth, \reeimpl) employed in \citet{du-etal-2021-grit,du-etal-2021-template} and \citet{huang-etal-2021-document} unfortunately departs from the stated metric \emph{definition} (\reedef) in two ways: (1) it eliminates the constraint on entity alignments and (2) it treats the template \emph{type} as an additional slot when reporting cross-slot averages. For maximally transparent comparisons to prior work, we report scores under both \reedef and \reeimpl, obtaining state-of-the-art results on \muc with each.

However, we argue that neither \reedef nor \reeimpl is consistent with historical evaluation of template extraction systems. \reedef\ errs in enforcing the entity alignment constraint: doing so effectively requires systems to perform coreference resolution, which is too strict and runs contrary to the original MUC-4 evaluation. By contrast, \reeimpl~also errs in treating the template type as just another slot: this elides the important distinction between the \emph{kind} of event being described and the participants in that event (\S\ref{sec:background}).

In the interest of clarity, we define a modified version of the CEAF-REE metric that avoids both pitfalls: it relaxes the entity alignment constraint and it does not include template type in cross-slot averages. We call this version CEAF-R\textbf{M}E, where ``M'' stands for \textit{mention} and emphasizes the focus on mention-level rather than entity-level (``E'') scoring. Intuitively, relaxing this constraint amounts to placing the burden of coreference resolution on the metric: if the scorer aligns two predicted mentions to the same reference entity, the mentions are implicitly deemed coreferent.

\begin{figure}[H]
    \centering
    \includegraphics[width=0.7\linewidth]{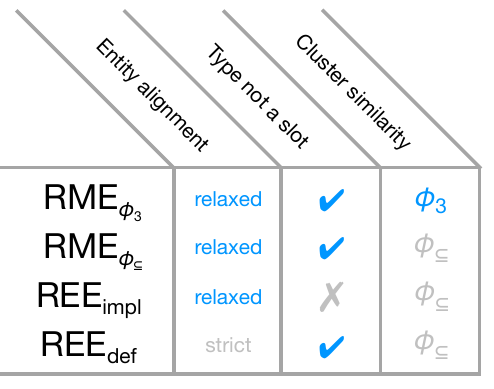}
    \caption{A comparison of the metrics discussed. Features in blue are ``desired'' for the evaluation of our task.}
    \label{fig:metric-analysis}
\end{figure}

Note that for a CEAF-family metric, the similarity function for entities $\phi(R, S)$ between the reference entity $R$ and the predicted $S$ is arbitrary \citep{luo-2005-coreference}. In \citet{du-etal-2021-grit},  \reeimpl uses $\phi_{\subseteq}(R, S) = \mathbf{1}[S \subseteq R]$. We argue that $\phi_{\subseteq}$ overly penalizes models for predicting incorrect mentions, as even a single incorrect mention reduces the score to 0. Instead, a better choice is $\phi_3(R, S) = \left| R \cap S \right|$ from \citet{luo-2005-coreference}: this computes a micro-average score of all mentions, and it adequately assigns partial credit to the overlap between the predicted mention set and the reference mention set.
See \autoref{fig:metric-analysis} for a succinct comparison among these variants.\footnote{~See \href{https://www.github.com/wanmok/iterx}{\tt github.com/wanmok/iterx} for implementations.} A more detailed discussion can be found in \autoref{app:evaluation}.

For \scirex, we report CEAF-RME (under both $\phi_3$ and $\phi_\subseteq$). For MUC-4, we report all metrics so that fair comparison with prior work can be made. For BETTER Granular, we use its official metrics, described in \autoref{app:evaluation}.

\vspace{-6.5mm}

\begin{table*}[t]
    \centering
    \adjustbox{max width=0.62\linewidth}{
    \begin{tabular}{l|ccccccccc}
    \toprule
         \bf Model (Encoder) & \multicolumn{3}{c}{\rmephi} & \multicolumn{3}{c}{\rmesubset}  \\
        \cmidrule(lr){2-4} \cmidrule(lr){5-7} 
        & P &  R & ${\rm F}_1$ & P & R & ${\rm F}_1$  \\
    \midrule
        \textsc{TempGen} (BART\textsubscript{base}) & 8.7 & 5.0 & 6.4 & 8.6 & 8.2 & 8.4   \\
        \textsc{TempGen} (BART\textsubscript{large}) & 19.9 & 5.0 & 8.0 & 8.9 & 22.3 & 12.7   \\
    \midrule 
        \iterx (BERT\textsubscript{base}) & 16.2 & 7.6 & 10.4 & 16.2 & 17.4 & 16.8   \\
        \iterx (${\rm BART}_{\rm base}^{\rm enc}$) & 15.0 & \bf 15.0 & 15.0 & 14.3 & 35.4 & 20.3  \\
        \iterx (${\rm T5}_{\rm large}^{\rm enc}$) & \bf 26.4 & 12.4 & \bf 16.9 & \bf 25.0 & \bf 40.6 & \bf 31.0   \\
    \bottomrule
    \end{tabular}
    }
    \caption{Results on \scirex.}
    \label{tab:exps-2}
\end{table*}

\begin{table*}[t]
    \centering
    \adjustbox{max width=1\linewidth}{
    \begin{tabular}{l|cccccccccccc}
    \toprule
         \bf Model (Encoder) & \multicolumn{3}{c}{\rmephi} & \multicolumn{3}{c}{\rmesubset} & \multicolumn{3}{c}{\reedef \textsuperscript{\dag}} & \multicolumn{3}{c}{\reeimpl} \\
        \cmidrule(lr){2-4} \cmidrule(lr){5-7} \cmidrule(lr){8-10}  \cmidrule(lr){11-13}
         & P & R & ${\rm F}_1$ & P & R & ${\rm F}_1$ & P & R & ${\rm F}_1$ & P & R & ${\rm F}_1$\\
    \midrule
        \textsc{TempGen} (BART\textsubscript{base})  & 54.2 & 15.8 & 24.5 & 58.3 & 31.0 & 40.5 & \it 53.1 & \it 29.5 & \it 37.9  & 55.7 & 40.0 & 46.4  \\
        \textsc{TempGen} (BART\textsubscript{large})  & \bf 55.8 & 18.9 & 28.3 & \bf 61.3 & 32.9 & 42.8 & \it 60.3 & \it 31.2 & \it 41.1 & \bf 63.7 & 37.4 & 47.2  \\
        GTT (BERT\textsubscript{base}) & 54.7 & 23.0 & 32.3 & 55.0 & 36.8 & 44.1 & \it 54.7 &\it  37.0 &\it  44.1 &  61.7 & 42.4 & 50.2  \\
    \midrule 
        \iterx (BERT\textsubscript{base}) & 41.3 & \bf 27.9 & 33.3 & 47.2 & \bf 45.0 & 46.1  & \it  41.3 &\it  45.3 & \it 43.2  & 52.3 & \bf 51.1 & 51.7 \\
        \iterx (${\rm BART}_{\rm base}^{\rm enc}$) & 39.2 & 24.8 & 30.4 & 44.8 & 40.1 & 42.3 & \it 35.4 & \it 20.3 & \it 39.2 & 49.8 & 45.7 & 47.6  \\
        \iterx (${\rm T5}_{\rm large}^{\rm enc}$) & 53.5 & 26.2 & \bf 35.2 & 55.8 & 42.4 & \bf 48.2 &\it  47.5 & \it 42.4 & \it  44.8 & 60.9 & 46.9 & \bf 53.0 \\
    \bottomrule
    \end{tabular}
    }
    \caption{Results on \muc. \small{\textsuperscript{\dag} Note that scores under \reedef are compared as if every mention forms a singleton entity. We made this assumption since neither prior work nor our model perform coreference resolution. Thus, \reedef is a somewhat inappropriate metric for these systems, but is included for completeness.}}
    \label{tab:exps-3}
\end{table*}

\subsection{Results}\label{subsec:exp-results}

\paragraph{\textsc{SciREX}}

For \textsc{TempGen}, we report models trained with BART\textsubscript{base} and BART\textsubscript{large}, where only BART\textsubscript{base} was used in \citet{huang-etal-2021-document}. While BART is an encoder-decoder architecture, \iterx uses only the encoder part, and thus requires about half the pretrained parameters that \textsc{TempGen} does.\footnote{~We add the superscript ``\textsuperscript{enc}'' to pretrained models to denote the use of the encoder only (decoder discarded): ${\rm T5}_{\rm large}^{\rm enc}$.} 

Even with far fewer parameters, \iterx outperforms the BART\textsubscript{large} baseline by a wide margin. Moreover, our best performing model under ${\rm T5}_{\rm large}^{\rm enc}$ \cite{RaffelSRLNMZLL20} achieves roughly $2\times$ the performance of \textsc{TempGen}\footnote{~Replacing BART with T5 in \textsc{TempGen} would have mandated destructive modifications to the pretrained architecture, and we therefore do not report results under this setting.} (see \autoref{tab:exps-2}).

\paragraph{MUC-4}
Under the most comparable setting, \iterx outperforms GTT under all metrics by 1--2\

\paragraph{BETTER Granular}

We report scores on the English-only Phase II BETTER Granular task using the official BETTER scoring metric in \autoref{tab:granular-exps}. Given the complexity of the Granular task, the accompanying difficulty of developing models to perform it, and the lack of existing work on Granular, we report scores only for \iterx under ${\rm T5}_{\rm large}^{\rm enc}$. We intend these to serve as a solid baseline against which future work may be measured.

\begin{table}[h]
    \centering
    \adjustbox{max width=1.\linewidth}{
    \begin{tabular}{ccccccc}
        \toprule
        \multicolumn{3}{c}{\bf Template} & \multicolumn{3}{c}{\bf Slot} & \bf Combined\\
        \cmidrule(lr){1-3} \cmidrule(lr){4-6}
        P &  R & ${\rm F}_1$ & P & R & ${\rm F}_1$ & \\
        \midrule
        89.7 & 74.5 & 81.4 & 41.0 & 33.5 & 36.9 & 30.0\\

        \bottomrule
    \end{tabular}
    }
    \caption{Results on the BETTER Granular dataset. Combined score is Template F1 $\times$ Slot F1.}
    \label{tab:granular-exps}
    \vspace{-3mm}
\end{table}

\section{Analysis}\label{sec:analysis}

We next conduct ablations to examine how specific aspects of \iterx's design affect learning. Here, we focus on \scirex as a case study, as it has the highest average templates per document of the three datasets, allowing us to best investigate the behavior of \iterx  over long action sequences.

Recall that the dynamic oracle specifies an expert policy $\pi^*$ (\autoref{eq:expert-policy}) from which expert actions $a^*$ are drawn. One design decision concerns the \emph{agent roll-out rate}, $\alpha$, which controls how often we draw from the expert policy vs. the agent policy when making updates. Another decision concerns how \emph{entropic} this policy distribution should be,  controlled by the temperature $\beta$. 

Both decisions reflect a trade-off between exploration and exploitation in the space of action sequences.

\paragraph{Agent Roll-out Rate $\boldsymbol{\alpha}$}

\begin{figure}[H]
    \centering
    \includegraphics[width=\columnwidth]{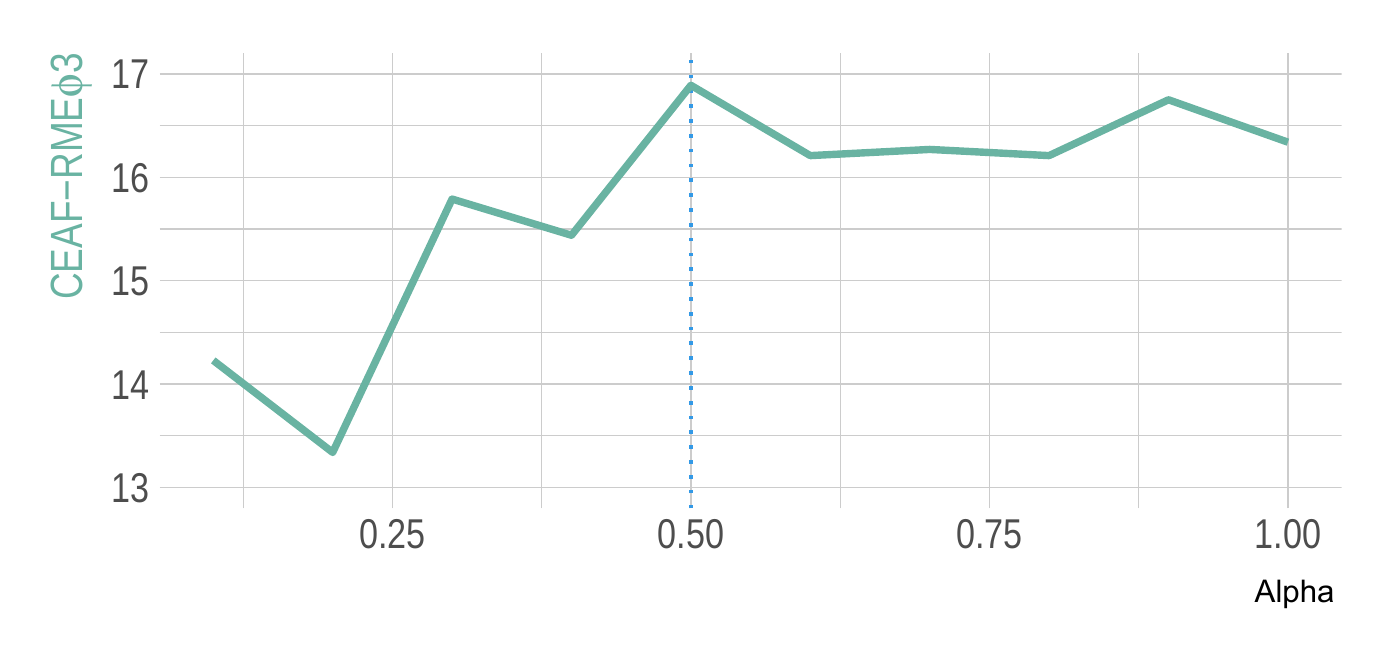}
    \caption{Performance changes on \rmephi with respect to $\alpha$, for which higher means higher probability of rolling out agent's policy for state update.}
    \label{fig:alpha-ree-change}
\end{figure}

We show how model performance changes as we increase the agent roll-out rate $\alpha \in [0, 1]$ in \autoref{fig:alpha-ree-change}, where $\alpha = 0$ to always following the expert policy, and $\alpha = 1$ corresponds to always updating based on the agent's own policy. The model performs poorly under low $\alpha$, but improves quickly as $\alpha$ increases, reaching a plateau past $\alpha \ge 0.5$. The results are intuitive, as relying more on the expert (lower $\alpha$) for learning would result in a fixed and deterministic set of states that may hinder the agent from visiting new states, which are often encountered at test time. With higher $\alpha$, the agent's behavior is more consistent between train and test time.

\paragraph{Temperature $\boldsymbol{\beta}$} We compare the following four settings for sampling from $\pi^*$, keeping $\alpha = 0$ to control for effects of policy mixing: 
\begin{itemize}\setlength\itemsep{-0.02em}
    \item \textsc{Fixed}: Select the next template in the document based on the order as is given in the dataset. In this case, $\beta$ does not come into play. This setting corresponds to the standard practice of using fixed template linearizations \citep{du-etal-2021-template, huang-etal-2021-document}.
    \item $\beta \to 0^{+}$ (\textsc{Argmax}): Select the template that maximizes the likelihood with the system-predicted distributions over slots. 
    
    \item $\beta = 1$ (\textsc{XEnt}): Sample a template according to the distribution defined by the cross entropy between references and predictions.
    \item $\beta \to \infty$ (\textsc{Uniform}): Sample a template uniformly from the correct template set.
\end{itemize}
Test set performance for each setting is shown in \autoref{tab:scirex-abl-beta}. 

The results for \rmesubset show a trade-off in precision and recall corresponding to the exploitation-exploration trade-off induced by $\beta$, with the higher $\beta$ (more exploration) of \textsc{XEnt} and \textsc{Uniform}, yielding higher recall. The trend for \rmephi is similar, though less pronounced. 

\begin{table}[h]
    \centering
    \adjustbox{max width=1.\linewidth}{
    \begin{tabular}{lcccccc}
    \toprule
        \bf Approach & \multicolumn{3}{c}{\bf \rmephi} & \multicolumn{3}{c}{\bf \rmesubset} \\
        \cmidrule(lr){2-4} \cmidrule(lr){5-7}
        & P &  R & ${\rm F}_1$ & P & R & ${\rm F}_1$ \\
    \midrule
        \textsc{Fixed} & 28.7 & 7.3 & 11.7 & 29.9 & 29.9 & 29.9 \\
    \midrule
        \textsc{Argmax} & 26.4 & 6.9 & 11.0 & 29.4 & 30.2 & 29.8 \\
        \textsc{XEnt} & 28.7 & 10.7 & 15.6 & 27.8 & 32.3 & 29.9 \\
        \textsc{Uniform} & 29.1 & 11.3 & 16.3 & 27.6 & 32.0 & 29.7 \\
    \bottomrule
    \end{tabular}
    }
    \caption{Results with different choices of temperature.}
    \label{tab:scirex-abl-beta}
\end{table}

Interestingly, while \rmesubset F\textsubscript{1} scores are consistent across settings, \rmephi F\textsubscript{1} scores are higher under higher temperature settings. To the extent that the more entropic settings conduce to higher template and mention recall, we would expect these settings to yield more partial-credit template alignments (under the $\phi_3$ similarity function) than non-random settings, which tend to focus on correct prediction of fewer templates --- thus potentially missing templates entirely and receiving no partial credit.

\section{Related Work}\label{sec:background}

\paragraph{Template Extraction}

The term \emph{template extraction} was originally proposed in the Message Understanding Conferences \citep[MUC; ][\emph{i.a.}]{sundheim-1991-overview} to describe the task of extracting templates from articles. Researchers later focused more heavily on sentence-level IE, especially after the release of the ACE 2005 dataset \citep{ace05}. But following renewed interest in document-level IE, researchers \citep[\emph{i.a.}]{du-etal-2021-template, huang-etal-2021-document, gantt2022event} have begun to revisit MUC and to develop new template extraction datasets (notably, BETTER Granular).

Traditionally, template extraction comprises two sub-tasks: \textit{template identification}, in which a system identifies and types all templates in a document, and \textit{slot filling} or \textit{role-filler entity extraction} (REE), in which the slots associated with each template are filled with extracted entities. Much recent work in this domain has turned away from the full task, focusing only on REE, which is tantamount to assuming that there is just a single \textit{aggregate} template per document \citep{patwardhan-riloff-2009-unified, huang-riloff-2011-peeling, huang-riloff-2012-modeling, du-etal-2021-grit, huang-etal-2021-document}.

\paragraph{Document-Level Relation Extraction}
Alongside template extraction, there has been considerable recent interest within IE in various challenging document-level relation extraction objectives, beyond the longstanding and dominant focus on coreference resolution. \textit{Argument linking} --- a generalization of semantic role labeling \citep[SRL;][]{gildea-jurafsky-2002-automatic} in which a predicate's extra-sentential arguments must also be labeled --- is one notable example, and has attracted recent attention through the RAMS \citep{ebner-etal-2020-multi} and WikiEvents \citep{li-etal-2021-document} benchmarks.\footnote{~Argument linking also goes by the names \emph{event argument extraction} and \emph{implicit semantic role labeling}, though these terms are not precisely equivalent.} Prior benchmarks on this task include SemEval 2010 Task 10 \citep{ruppenhofer-etal-2010-semeval}, Beyond Nombank \citep{gerber-chai-2010-beyond}, ONV5 \citep{moor-etal-2013-predicate}, and Multi-sentence AMR \citep{ogorman-etal-2018-amr}. A separate line of work has concentrated on general \textit{N-ary relation extraction} challenge tasks, in which entities participating in the same relation may be scattered widely throughout a document. Beyond \scirex, PubMed \citep{quirk-poon-2017-distant, peng-etal-2017-cross} and DocRED \citep{yao-etal-2019-docred} are two other prominent benchmarks in this area.

\paragraph{Imitation Learning}
Our approach casts the problem of generalized template extraction as a Markov decision process. SEARN \cite{DaumeLM09} and other related work \cite[][i.a.]{RossGB11,VenkatramanHB15,ChangKADL15} have considered structured prediction under a reinforcement learning setting. Notably, in dependency parsing, \citet{goldberg-nivre-2012-dynamic} proposed the use of a \emph{dynamic oracle} to guide an agent toward the correct parse (see \S\ref{subsec:training}).

We also employ direct policy learning for optimization of the template extraction MDP, thus reducing the problem to one of supervised sequence learning that is amenable to gradient descent. Such treatment is reminiscent of other similar techniques in NLP. Scheduled sampling \cite{BengioVJS15}, for instance, trains a sequence generator with an expert policy consisting of a mixture of the predicted token and the gold token. Relatedly, Levenshtein Transformers \cite{GuWZ19} learn to edit a sequence by imitating an expert policy based on the Levenshtein edit distance.

\section{Conclusion}\label{sec:conclusion}
We have presented \iterx, a new model for generalized template extraction that iteratively generates templates via a Markov decision process. \iterx demonstrates state-of-the-art performance on two benchmarks in this domain --- 4-ary relation extraction on \scirex and template extraction on \muc\ --- and establishes a strong baseline on a third benchmark, BETTER Granular. In our experiments, we have also shown that imitation learning is a viable paradigm for these problems. We hope that our findings encourage future work to confront the challenge of dealing with documents that describe \textit{multiple} complex events and relations head-on, rather than veiling this difficulty behind simplified task formulations.

\section{Limitations}\label{sec:limitations}

Although we believe our iterative extraction paradigm to be promising, we acknowledge that this work is not without limitations. First, \iterx features a significant number of hyperparameters. We found that these generally required some effort to tune for specific datasets, and that there was no single configuration that was uniformly the best across domains. We showed the impact of manipulating some of these hyperparameters in \S\ref{sec:analysis}. Second, our \iterx implementation iterates over all template types in the template ontology during training and inference, which means that runtime grows linearly in the number of template types. While our framework could in principle support template \emph{type} prediction as well (which would reduce this to $O(1)$), it does not do so in practice, and hence runtime may be long for large ontologies. However, we again stress that actual template ontologies tend to be small.

\section*{Acknowledgments}

We thank Nathaniel Weir, Elias Stengel-Eskin, and Patrick Xia for helpful comments and feedback. This work was supported in part by DARPA AIDA (FA8750-18-2-0015) and IARPA BETTER (2019-19051600005). The views and conclusions contained in this work are those of the authors and should not be interpreted as necessarily representing the official policies, either expressed or implied, or endorsements of DARPA, IARPA, or the U.S. Government. The U.S. Government is authorized to reproduce and distribute reprints for governmental purposes notwithstanding any copyright annotation therein.

\bibliography{anthology,custom}
\bibliographystyle{acl_natbib}

\appendix

\section{Terminology}\label{app:terminology}
Information Extraction is rife with vague and competing terms for similar concepts, and we recognize some hazard in introducing \textit{generalized template extraction} (GTE) into this landscape. To head off possible confusion, we highlight two important differences between this problem and the well established problem of \textit{event extraction} (EE).

First, EE requires identifying \textit{lexical} event triggers, whereas GTE does not, as template instances do not necessarily have one specific lexical anchor. A document describing a terrorist attack may only \textit{explicitly} describe a series of bombings, and a document describing an epidemic may only \textit{explicitly} state that thousands of people have contracted a particular disease. This property holds of all three datasets we focus on, and can be seen in both \autoref{fig:multi_template} and \autoref{fig:muc-better-egs}. Template anchors are not annotated either for \muc or for \scirex. And while they are annotated for BETTER Granular, they do not factor into scoring. This contrasts with major EE datasets, such as ACE or PropBank, for which typed lexical triggers \textit{must} be extracted.

Second, we take GTE to be a fundamentally document-level task: templates concern events described over an entire document. In practice, EE has historically referred to extraction of predicate-argument structures within a single sentence. One could conceivably argue that this usage has begun to change with the recent interest in argument linking datasets like RAMS \citep{ebner-etal-2020-multi} and WikiEvents \citep{li-etal-2021-document}, in which arguments may appear in different sentences from the one containing their predicate. Even so, these cross-sentence arguments are still arguments \textit{of a particular predicate}, \textit{in a particular sentence}. Moreover, the overwhelming majority of arguments in these datasets are sentence-local \citep{ebner-etal-2020-multi}. As emphasized above, templates are not necessarily anchored to particular lexical items. For this reason, they also do not necessarily exhibit the level of locality one finds in EE.

These differences are what motivate the use of CEAF-REE as an evaluation metric, in contrast to the precision, recall, and F1 scores for events and arguments that are typically reported for EE. In brief, it simply is not possible to compute these for GTE in the same way as they are computed for EE. We elaborate on this point in Appendix \ref{app:evaluation}.

\section{Model Training and Hyperparameters}\label{app:architecture}
We implemented our models in PyTorch \cite{Paszke2019} and AllenNLP \cite{gardner2018allennlp}. We trained all our models with a single NVIDIA RTX6000 GPU. For all experiments that reproduce prior works, we trained models until full convergence under the patience settings provided in the publicly released code. For all \iterx models, we trained and tuned hyperparameters under our grid's limit of 24 hours per run, with which we were able to obtain solid performance on all datasets. We performed hyperparameter search manually and report the best performing hyperparameters and the bounds we searched in \autoref{tab:app-scirex-hyperparams}, \autoref{tab:app-muc4-hyperparams}, and \autoref{tab:app-granular-hyperparams}.

\begin{table}[h]
    \centering
    \adjustbox{max width=\linewidth}{
    \begin{tabular}{lcc}
        \toprule
        Name & Best & Search Bounds \\
        \midrule
        Encoder & ${\rm T5}_{\rm large}^{\rm enc}$ & $\{ {\rm T5}_{\rm base}^{\rm enc}, {\rm T5}_{\rm large}^{\rm enc}, {\rm BART}_{\rm base}^{\rm enc}, {\rm BART}_{\rm large}^{\rm enc}, {\rm BERT}_{\rm base}\}$ \\
        LR & $3\times 10^{-5}$ & $\{1\times 10^{-5}, 3\times 10^{-5}, 5\times 10^{-5} \}$ \\
        Encoder LR & $1\times 10^{-5}$ & $1\times 10^{-5}$ \\
        $\alpha$ & 0.5 & $[0, 1]$ \\
        $\beta$ & 1.0 & $\{0, 1.0, \infty\}$ \\
        $\gamma$ & 1.0 & \{1.0\} \\
        $\pi$ Type & Joint & \{Independent, Joint\} \\
        Max \#Iteration & 10 & $\{10, 30\}$ \\
        Training Spans & Gold & \{Gold, Upstream\} \\
        \midrule 
        \midrule 
        Avg. training time & \multicolumn{2}{c}{3 hrs} \\
        Validation Metric & \multicolumn{2}{c}{CEAF-RME} \\
        \# of parameters\textsuperscript{*} & \multicolumn{2}{c}{$\sim$362 million} \\
        \bottomrule
    \end{tabular}
    }
    \caption{Hyperparameters and other reproducibility information for \scirex. ``LR'' denotes learning rate, ``$\pi$ Type'' indicates which policy network architecture is used (see \S\ref{subsec:policy}), ``Max \#Iteration'' sets the maximum number of iterations that the model is allowed to perform, and ``Training Spans'' determines whether the training spans come from gold annotations or the intersection of gold spans and those predicted by the span finding module. \textsuperscript{*}The number of trainable parameters include the parameters from the (best) encoder.}
    \label{tab:app-scirex-hyperparams}
\end{table}

\begin{table}[h]
    \centering
    \adjustbox{max width=\linewidth}{
    \begin{tabular}{lcc}
        \toprule
        Name & Best & Search Bounds \\
        \midrule
        Encoder & ${\rm T5}_{\rm large}^{\rm enc}$ & $\{ {\rm T5}_{\rm base}^{\rm enc}, {\rm T5}_{\rm large}^{\rm enc}, {\rm BART}_{\rm base}^{\rm enc}, {\rm BART}_{\rm large}^{\rm enc}, {\rm BERT}_{\rm base}\}$ \\
        LR & $3\times 10^{-5}$ & $\{1\times 10^{-5}, 3\times 10^{-5}, 5\times 10^{-5} \}$ \\
        Encoder LR & $1\times 10^{-5}$ & $1\times 10^{-5}$ \\
        $\alpha$ & 0.6 & $[0, 1]$ \\
        $\beta$ & 1.0 & $\{0, 1.0, \infty\}$ \\
        $\gamma$ & 1.0 & \{1.0\} \\
        $\pi$ Type & Independent & \{Independent, Joint\} \\
        Max \#Iteration & 14 & $\{14\}$ \\
        Training Spans & Upstream & \{Gold, Upstream\} \\
        \midrule 
        \midrule 
        Avg. training time & \multicolumn{2}{c}{20 hrs} \\
        Validation Metric & \multicolumn{2}{c}{CEAF-RME} \\
        \# of parameters & \multicolumn{2}{c}{$\sim$379 million} \\
        \bottomrule
    \end{tabular}
    }
    \caption{Hyperparameters and other reproducibility information  for \muc. }
    \label{tab:app-muc4-hyperparams}
\end{table}

\begin{table}[h]
    \centering
    \adjustbox{max width=\linewidth}{
    \begin{tabular}{lcc}
        \toprule
        Name & Best & Search Bounds \\
        \midrule
        Encoder & ${\rm T5}_{\rm large}^{\rm enc}$ & $\{ {\rm T5}_{\rm base}^{\rm enc}, {\rm T5}_{\rm large}^{\rm enc}, {\rm BART}_{\rm base}^{\rm enc}, {\rm BART}_{\rm large}^{\rm enc}\}$ \\
        Optimizer & AdamW & \{AdamW\} \\
        LR & $3\times 10^{-5}$ & $\{1\times 10^{-5}, 3\times 10^{-5}, 5\times 10^{-5} \}$ \\
        Encoder LR & $1\times 10^{-5}$ & $1\times 10^{-5}$ \\
        $\alpha$ & 048 & $[0, 1]$ \\
        $\beta$ & $\infty$ & $\{0, 1, \infty\}$ \\
        $\gamma$ & 1.0 & \{1.0\} \\
        $\pi$ Type & Joint & \{Independent, Joint\} \\
        Max \#Iteration & 10 & $\{ 10, 30\}$ \\
        Training Spans & Gold & \{Gold, Upstream\} \\
        \midrule 
        \midrule 
        Avg. training time & \multicolumn{2}{c}{24 hrs} \\
        Validation Metric & \multicolumn{2}{c}{BETTER combined score} \\
        \# of parameters & \multicolumn{2}{c}{$\sim$591 million} \\
        \bottomrule
    \end{tabular}
    }
    \caption{Hyperparameters and other reproducibility information  for Granular.}
    \label{tab:app-granular-hyperparams}
\end{table}

\section{Dataset Details}\label{app:datasets}

\subsection{MUC-4}\label{subsec:muc4}
The \muc\ dataset features a total of 1,700 English documents (1,300 for train and 200 each for dev and test) concerning geopolitical conflict and terorrism in South America. Documents are annotated with templates of one of six kinds --- \texttt{Attack}, \texttt{Arson}, \texttt{Bombing}, \texttt{Murder}, \texttt{Robbery}, and \texttt{ForcedWorkStoppage} --- and may have multiple templates (often of the same type) or no templates at all. All templates contain the same slots. While the original data contains numerous slots, it has become standard practice to evaluate systems on just five of these (apart from the slot for the template's type), all of which take entity-valued fillers: \texttt{Perpetrator(Individual)}, \texttt{Perpetrator(Organization)}, \texttt{Victim}, \texttt{Weapon}, and \texttt{Target}.

\subsection{BETTER Granular}\label{subsec:better}
The BETTER Granular dataset contains documents spanning a number of domains, and, like \muc, focuses on six types of complex event, though covering different topics: protests, epidemics, natural disasters, acts of terrorism, incidents of corruption, and (human) migrations. However, Granular is substantially more difficult than \muc\ in several ways. First, each template type is associated with a \emph{distinct} set of slots. Second, only some of the slots take entities as fillers, whereas others take events, boolean values, or one of a fixed number of strings. Finally, the formal evaluation setting for Granular --- which we do not adopt in this paper --- is zero-shot and cross-lingual: systems trained only on English documents are evaluated exclusively on documents in a different target language.\footnote{Systems are permitted to use machine-translated versions of this documents, but gold data in the target language is prohibited.} The data used in our experiments is English-only and comprises the ``train,'' ``analysis,'' and ``devtest'' splits from Phase II of the BETTER program, for which the target language is Farsi.

\subsection{\scirex}\label{subsec:scirex}
The 4-ary \scirex relation extraction task seeks to idenfity entity 4-tuples that describe a \texttt{metric} used to evaluate a \texttt{method} applied to an ML \texttt{task} as realized by a specific \texttt{dataset} --- e.g. (\textit{span F1}, \textit{BERT}, \textit{SRL}, \textit{ACE 2005}). The challenge of \scirex lies not only in these pieces of information tending to be widely dispersed throughout an article, but also in the fact that only tuples describing \textit{novel} work presented in the paper (and not merely \textit{cited} work) are labeled as gold examples. Following \citet{huang-etal-2021-document}, we frame this as a template extraction task, treating each 4-tuple as a template with four slots.

\section{Model Evaluation Details}\label{app:evaluation}
A key consideration that arises in evaluating generalized template extraction is the need to align predicted and reference templates: a given predicted template may be reasonably similar to multiple different reference templates, and one must decide on a \textit{single} template to use as the reference for each predicted one. Generalized template extraction is similar in this respect to coreference resolution, in which predicted \textit{entities} may (partially) match multiple reference entities, and one must determine a ground truth alignment. Importantly, this consideration also renders metrics that are traditionally reported for \textit{event extraction} --- namely, event and argument precision, recall, and $\rm F_1$ --- inappropriate. This is because event extraction is fundamentally a span labeling problem, and the identity of the appropriate reference span is always clear for a given predicted span: either a reference span with the same boundary and type exists or it does not.\footnote{In the case of argument spans, one may additionally require the associated trigger spans to match as well.} By contrast, the mapping from prediction to reference for templates is only this transparent in cases of perfectly accurate predictions.

All the evaluation metrics presented in this appendix are, at base, minimal extensions of precision, recall, and $\rm F_1$ to cases where template alignments are both necessary and non-trivial. For CEAF-REE in particular, the various versions of the metric that we discuss (\rmephi, \rmesubset, \reedef, and \reeimpl) merely reflect differences in how this alignment should be performed and whether the template type should be treated in the same way as slot types for reporting purposes.

\subsection{MUC-4}\label{app:eval-muc4}
MUC-4 evaluation presents a special challenge, owing to its long and complicated history, and to terminological confusion.\footnote{Early writing on MUC-4 used the term \textit{entity} to refer to what the IE community would now call a \textit{mention}. We suspect this is the source of a great deal of confusion.} Here, we discuss CEAF-REE \citep{du-etal-2021-grit}, the current standard metric for MUC-4 evaluation. We begin with definitions, following with a discussion of some of its problems, and conclude with an extended presentation of our CEAF-RME variant, introduced in \S\ref{sec:experiments}.

\subsubsection{CEAF and CEAF-REE: Definitions}
The CEAF-REE metric, introduced by \citet{du-etal-2021-grit}, has since been adopted as the standard evaluation metric for MUC-4 \citep{du-etal-2021-template, huang-etal-2021-document}. To our knowledge, no official scoring script has ever been released for \muc, although the metrics used as part of the original evaluation are described in detail in \citet{chincnor-1992-muc4}. CEAF-REE does not attempt to implement these original metrics, but is rather a lightly adapted version of the widely used CEAF metric for coreference resolution, proposed in \citet{luo-2005-coreference}.\footnote{~\citeauthor{luo-2005-coreference}'s motivations for proposing CEAF actually derive in large part from observed shortcomings with the original MUC-4 F\textsubscript{1} score. See \citet{luo-2005-coreference} for details.} CEAF computes an alignment between reference ($\mathcal{R}$) and system-predicted ($\mathcal{S}$) entities, with each entity represented by a set of coreferent mentions, and with the constraint that each predicted entity is aligned to \emph{at most} one reference entity. This is treated as a maximum bipartite matching problem, in which one seeks the alignment that maximizes the sum of an entity-level similarity function $\phi(R,S)$ over all aligned entities $R \in \mathcal{R}$ and $S \in \mathcal{S}$ within a document. In principle, CEAF is agnostic to the choice of $\phi$, though it is generally desirable that $\phi(R,S) = 0$ when $\nexists x \in S$ such that $x \in R$ and that $\phi(R,S) = 1$ when $R = S$, for reasons described in \citet{luo-2005-coreference}. In practice, the $\phi_4$ similarity function is most commonly used, defined as the Dice coefficient (or $\rm F_1$ score) between $R$ and $S$:
\begin{equation}
    \phi_4(R, S) := \frac{2|R \cap S|}{|R| + |S|}.
\end{equation}
Another possible version is $\phi_3$:
\begin{equation}
    \phi_3(R, S) = |R \cap S|.
\end{equation}

Given this $\phi$ similarity function and the maximal match $g^*$ between entities, the final precision and recall are computed as follows:
\begin{align}
    p &= \frac{{\displaystyle\sum}_{(R,S) \in g^*}~\phi(R, S)}{{\displaystyle\sum}_S~ \phi(S, S)}; \\
    r &= \frac{{\displaystyle\sum}_{(R,S) \in g^*}~\phi(R, S)}{{\displaystyle\sum}_R~ \phi(R, R)}.
\end{align}

Here we see that $\phi_4$ computes a version of \emph{macro-average} over entities, whereas $\phi_3$ computes a \emph{micro-average}.

The CEAF that uses $\phi_4$ is sensibly denoted $\text{CEAF}_{\phi_4}$ in coreference resolution. \reeimpl differs from $\text{CEAF}_{\phi_4}$ in the following ways:
\begin{itemize}
    \item All entities are aligned \textit{within role, conditioned on matching template type}. E.g. only predicted entities for the \texttt{Victim} slot in \texttt{Bombing} templates would be considered valid candidates for alignment with entities filling the \texttt{Victim} slot in reference \texttt{Bombing} templates.
    \item A binary similarity function $\phi_{\subseteq}$ is used, defined as follows:
    \begin{equation}\phi_{\subseteq}(R,S) := 
    \begin{cases}
    1, \quad\text{if } S \subseteq R \\
    0, \quad\text{otherwise}
    \end{cases}
    \end{equation}
    This function $\phi_\subseteq$ says that a model receives full credit (1.0) for a predicted entity if and only if its mentions form a subset of those in the reference entity. If even one incorrect mention is included, the model receives a score of zero for that entity.
\end{itemize}

\subsubsection{CEAF-REE: Problems and Solutions}
Our principal concerns with CEAF-REE lie with how it has so far been reported and implemented, and with challenges in extending it to the full template extraction task, in which multiple templates of the same type may be present in a document. We elaborate on two issues discussed briefly in \S\ref{sec:experiments} and also introduce a third.

First, previous work that reports CEAF-REE treats the template type merely as another slot, with template type labels treated as special kinds of ``entities'' that may fill this slot. This is not necessarily problematic in itself: template type-level metrics are valuable for evaluating system performance. However, it is problematic when reporting (micro or macro) average CEAF-REE figures \textit{across} slots, as these works do. This is because incorporating the scores for template type into the average elides the distinction between roles (slots) and the kind of event being described (the template type). Moreover, the alignment between slot-filling entities is also \textit{already} conditioned on a match between the template types. There are thus two distinct ways in which information about a system's predictive ability with respect to template type end up in a slot-level average CEAF-REE score. This results in reported values that are very difficult to interpret, and potentially misleading to the extent that these features of CEAF-REE implementations are not made apparent in writing.

\begin{figure}
    \centering
    \includegraphics[width=0.95\linewidth]{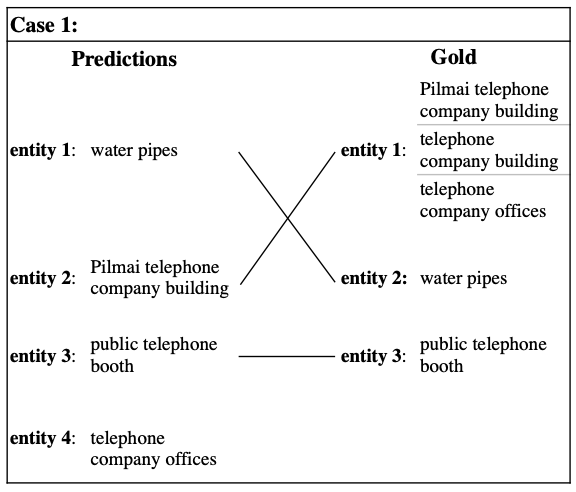}
    \caption{An example alignment between predicted and reference entities from \citet{du-etal-2021-grit}. In past implementations of CEAF-REE, this alignment would receive full credit, rather than being penalized for precision ($P = 0.75$).}
    \label{fig:ceaf_ree_ex}
\end{figure}

Second, the constraint that at most one predicted entity be aligned to each reference entity --- stipulated in the metric definition (\reedef) --- is not enforced in the implementation (\reeimpl). Practically, this means that the alignment shown in \autoref{fig:ceaf_ree_ex} would receive full credit, whereas it ought to receive a precision score of only 0.75, as \citet{du-etal-2021-grit} describe. As we argue in \S\ref{sec:experiments}, we believe this constraint to be overly strict. But this point aside, the discrepancy between definition and implementation is problematic in itself.

Third, full template extraction introduces a \textit{second} maximum bipartite matching problem, which requires aligning predicted and reference \textit{templates} of the same type, and which CEAF-REE (either \reedef or \reeimpl) is not natively equipped to handle, given that it operates at the level of slots. \citet{du-etal-2021-template} reports CEAF-REE for GTT under an optimal template alignment, but this is obtained via brute-force, enumerating and evaluating every possible alignment, including those between templates of different types. The similarity function, (call it $\phi_{\textsc{temp}}(T_R, T_S)$) that they use for template alignment is \emph{itself} the cross-slot average \reeimpl score for predicted template $T_S$ and reference template $T_R$. This brute-force template alignment, in conjunction with the two-level maximum bipartite matching problem, results in prohibitively long scorer execution times in cases where there are even a modest number of predicted or reference templates of the same type.\footnote{Only \reedef requires solving a two-level maximum-bipartite matching problem. Since \reeimpl does not enforce the entity alignment constraint, these alignments will not necessarily be bipartite.}

In addition to our implementation of CEAF-RME (see below), we also present the first \emph{correct} implementation of \reedef that fully addresses the first two points above: template types are no longer treated as additional slots and the entity-level alignment constraint is enforced. On the third point, our implementation efficiently computes optimal template alignments using the Kuhn-Munkres algorithm \cite{kuhn1955hungarian,munkres1957algorithms}. However, even with this efficient implementation, solving the two-level maximum bipartite matching problems is still computationally intensive.

\subsubsection{Coreference and CEAF-RME}
As CEAF was designed for coreference, it is unsurprising that coreference considerations introduce a further wrinkle for CEAF-REE. None of the three models described in this work (including \iterx) performs entity coreference resolution. This clearly presents a problem because \reedef is an entity-level metric. One way to score these models is simply to treat each extracted mention as a singleton entity and use \reedef exactly as defined, and we report these scores in the main text for MUC-4. However, reporting \emph{only} \reedef would be undesirable for several reasons:
\begin{itemize}
    \item It would render our results incomparable to past work, which reports only \reeimpl.
    \item It would put our work at odds with the overwhelming majority of the template extraction literature, where evaluation criteria focus on string matching between predicted and reference \textit{mentions}. (The original \muc evaluation only required systems to extract a single representative mention for each entity --- not to identify all such mentions.)
    \item The constraint that at most one predicted entity be aligned to a given reference entity would yield punishingly low scores for systems that are highly effective at extracting relevant spans, but that simply do not perform the additional step of coreference. 
\end{itemize}

For these reasons, we disfavor a template extraction metric that \textit{requires} template extraction systems to do coreference. These considerations motivate our introduction of CEAF-R\textbf{M}E (\textit{role-filler mention extraction}) --- that makes a minimal modification to \reedef to address (1) and (2) above. CEAF-RME treats system-predicted mentions as singleton entities, but \textit{deliberately} relaxes the alignment constraint, potentially allowing multiple predicted singletons to map to the same reference entity, effectively pushing the burden of coreference into the metric. We believe CEAF-RME is consistent with what template extraction research has \textit{in fact} historically cared about (identifying \textit{mentions} that fill some slot) while correcting implementation problems with CEAF-REE that produce misleading results.

The micro-average \rmephi results that we report on MUC-4 in the main body of the paper are micro-average CEAF-RME scores \emph{under an optimal template alignment} (using CEAF-RME as the template similarity function), which is efficiently obtained using the Kuhn-Munkres algorithm.

We additionally include a version of CEAF-RME that uses $\phi_\subseteq$ (\rmesubset) for parallel comparison against \reeimpl. Recall that \reeimpl is essentially \rmesubset with the template type included as an additional slot. We reiterate that \rmephi is the more appropriate metric since it can award partial credit for predicted entities whose mentions overlap imperfectly with those in the reference, where \rmesubset assigns zero credit in such cases.

\subsection{\scirex}
We use the same \rmephi implementations for scoring \scirex as we use for MUC-4. Full evaluation using the original \scirex scoring script requires systems to perform coreference resolution, which makes it similarly inappropriate to \reedef for evaluating the systems presented in this work, none of which feature a coreference module. The \rmephi and \rmesubset results presented in the main text together give a clearer picture of these models' ability to extract relevant mentions (short of clustering them) than would a coreference-based metric. We simply treat the \scirex\ 4-tuples as 4-slot templates, following \citet{huang-etal-2021-document}. 

\subsection{BETTER Granular}\label{app:eval-better}
Evaluation for the BETTER Granular task bears some core similarity to \reedef in that relies on obtaining the alignment between system and reference templates that maximizes some similarity function that decomposes over slot fillers. And just as with (our corrected implementation of) \reedef, this is achieved via the Kuhn-Munkres algorithm. However, Granular scoring differs from \reedef in four key respects. First, the overall system score --- referred to as the \textit{combined score} --- incorporates both a slot-level F1 score \textit{and} a template-level F1 score:
\begin{equation}
    \text{CombinedScore} := \text{TypeF1} \times \text{SlotF1} \nonumber
\end{equation}
Only exact matches between system and reference templates types are awarded credit. It is worth noting that because this score does not decompose over template pairs, it cannot be optimized directly using Kuhn-Munkres. In practice, what is optimized is \emph{response gain} --- the number of correct slot fillers minus the number of incorrect ones --- which provably yields alignments that optimize the combined score within a probabilistic error bound.

The remaining three key differences relate to the calculation of the slot-level F1. For one, Granular slots are not exclusively entity-valued, but may also be event-, (mixed) event-and-entity-, boolean-, and (categorical) string-valued, and different similarity functions must be employed in these different cases. For another, where CEAF-REE defines mentions by their string representation, the Granular score defines mentions based on document offsets. Finally, Granular also requires extraction of temporal and irrealis information for slots, and this in turn impacts the SlotF1 score.

Borrowing terminology from the discussion of MUC-4 above, we describe below how $\phi(R,S)$ is calculated for some generic reference slot filler $R$ and system-predicted slot filler $S$ for slots of different types.

\paragraph{Boolean and Categorical Values}
For boolean- and categorical-string valued slots (i.e., slots taking on one of a predefined set of values). $\phi(R,S) = 1$ if there is an exact match between the system and reference fillers and is 0 otherwise.

\paragraph{Entities}
Unique among the three tasks discussed in this paper, Granular features an explicit preference for \textit{informative} arguments in its scoring structure. In particular, (proper) \textit{name} mentions of an entity are worth more than \textit{nominal} mentions, which in turn are worth more than \textit{pronominal} ones.\footnote{This is precisely the hierarchy described for the \textit{informative argument extraction} task in \citet{li-etal-2021-document}.} Thus, if Barack Obama were represented by the reference entity $\{\textit{Obama}, \textit{the former President}, \textit{he}\}$, full credit would be awarded for returning only the mention \textit{Obama}, less credit for \textit{the former President}, and still less for \textit{he}. Exact point values depend on the mentions present in the reference entity:
\begin{itemize}
    \item Correct name mentions always receive full credit ($\phi(R,S) = 1$)
    \item Correct nominal mentions receive half-credit ($\phi(R,S) = 0.5$) if the reference entity additionally contains a name mention, and receive full credit otherwise.
    \item Correct pronominal mentions receive quarter-credit ($\phi(R,S) = 0.25$) if the reference entity additionally contains \textit{both} a name and a nominal mention, and half-credit if only a nominal mention is featured. (Note that entities will never feature only pronominal mentions.)
\end{itemize}

\paragraph{Events}
Some Granular slots require \emph{events} as fillers. Like entities, events are represented as sets of mentions (event \textit{anchors} or \textit{triggers}). Unlike entities, there is no informativity hierarchy for events. Furthermore, while event coreference is not a part of the Granular task, annotations for event coreference are nonetheless provided for scoring purposes: $\phi(R,S) = 1.0$ iff $S$ contains only mentions belonging to events in the set of gold coreferent events $R$, and is 0 otherwise, akin to $\phi_{\textsc{REE}}$.

\paragraph{Mixed Entities and Events}
Some slots may take a mix of events and entities as fillers. Since systems must indicate whether predicted mention clusters are entity- or event-denoting, the same similarity criteria for events and entities as described above are used to compute $\phi$ for events and entities that fill these slots.

\paragraph{Temporal and Irrealis Information}
One of the features of Granular that makes it decidedly more difficult than either MUC-4 or \scirex\ is the requirement to extract information relating to the time and irrealis status of an event when such information is available in the document. This information is encapsulated in special \texttt{time-attachments} and \texttt{irrealis} fields associated with each slot-filling entity or event. The former is given as a set of temporal expressions that describe the time at or during which the filler satisfied the role denoted by the slot (e.g. when individuals filling the \texttt{tested-count} slot in the Epidemic template were tested for the disease). The latter is given as one of a set of strings that describe whether or how the filler satisfied the role denoted by the slot: \texttt{counterfactual}, \texttt{hypothetical}, \texttt{future}, \texttt{unconfirmed}, \texttt{unspecified}, and \texttt{non-occurrence}. \texttt{time-attachments} and \texttt{irrealis} are each worth 0.25 points, where exact matches are required for full credit on either and where zero points are awarded otherwise. For slots for which \texttt{time-attachments} and \texttt{irrealis} are required, the value of $\phi$ appropriate to its filler type is scaled by 0.5 such that the maximum overall score $\phi(R,S)$ for a given filler --- factoring in \texttt{time-attachments}, \texttt{irrealis}, \textit{and} event or entity similarity --- is 1.

\end{document}